\documentclass[10pt,twocolumn,letterpaper]{article}

\usepackage{iccv}
\usepackage{times}
\usepackage{epsfig}
\usepackage{graphicx}
\usepackage{amsmath}
\usepackage{amssymb}
\usepackage{xcolor}
\usepackage{caption}
\usepackage{booktabs}
\usepackage{multirow}
\usepackage{soul}

\usepackage[breaklinks=true,bookmarks=false,pagebackref,colorlinks]{hyperref}

\iccvfinalcopy 

\def\iccvPaperID{9476} 

\ificcvfinal\fi

\newcommand\blfootnote[1]{%
  \begingroup
  \renewcommand\thefootnote{}\footnote{#1}%
  \addtocounter{footnote}{-1}%
  \endgroup
}

\newcommand{\xx}[0]{\mathbf{x}}
\newcommand{\cc}[0]{\mathbf{c}}
\newcommand{\kk}[0]{\mathbf{k}}
\newcommand{\vv}[0]{\mathbf{v}}

\newcommand{\tttt}[0]{\mathbf{w}}
\newcommand{\I}[0]{\mathbf{I}}
\newcommand{\D}[0]{D_{\theta}}
\newcommand{\K}[0]{\mathbf{K}}
\newcommand{\Q}[0]{\mathbf{Q}}
\newcommand{\V}[0]{\mathbf{V}}
\newcommand{\W}[0]{\mathbf{W}}
\newcommand{\M}[0]{\mathbf{M}}
\newcommand{\OO}[0]{\mathbf{O}}

\begin{document}

\title{Editing Implicit Assumptions in Text-to-Image Diffusion Models}

\author{Hadas Orgad$^*$ \qquad Bahjat Kawar$^*$ \qquad Yonatan Belinkov$^\dagger$ \\
Computer Science Department, Technion, Israel\\
{\tt\small \{orgad.hadas@cs., bahjat.kawar@cs., belinkov@\}technion.ac.il}
}

\twocolumn[{%
\renewcommand\twocolumn[2][]{#1}%
\maketitle%
\vspace{-0.4cm}	
\centering \centering
\includegraphics[width=\textwidth]{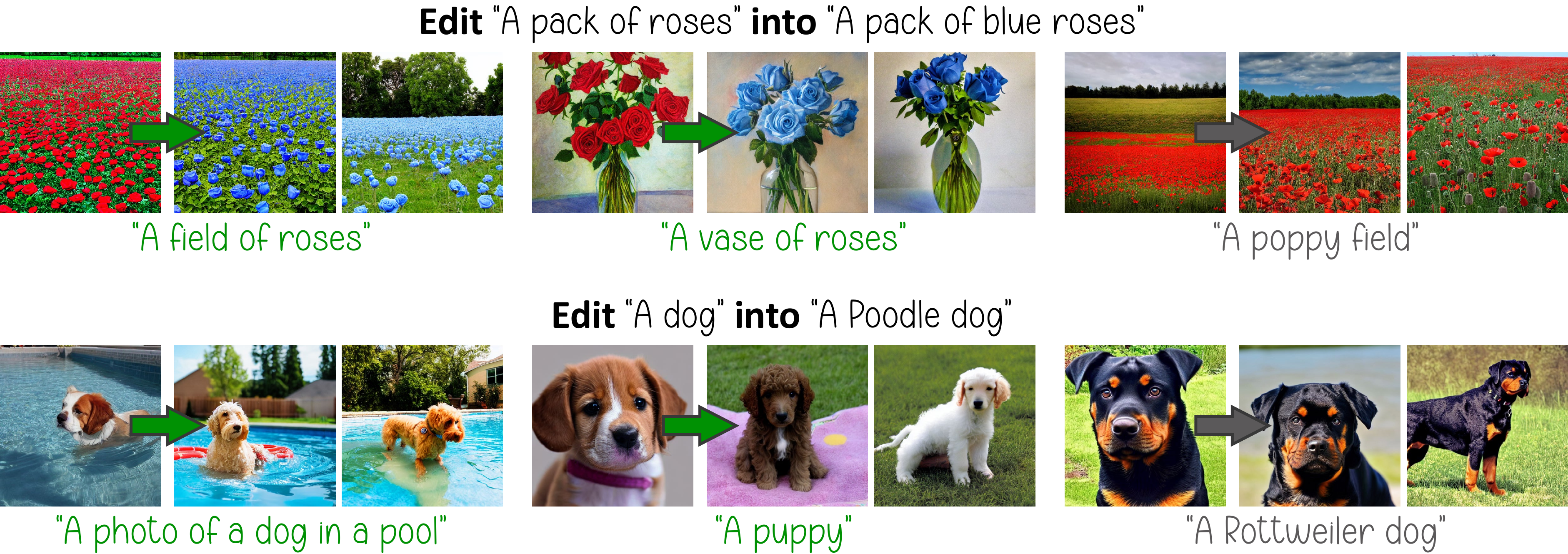}
\captionsetup[figure]{aboveskip=0.2cm}
\captionof{figure}
{TIME edits implicit assumptions in a model (\textit{e.g.}, roses are red). As a result, related prompts (green) change their behavior, while unrelated ones (gray) do not.
For example, after model editing, the roses in ``A field of roses'' become blue.
}
\vspace*{0.2cm}
\label{fig:headline}%
}] 
\blfootnote{$^*$ Equal contribution.}
\blfootnote{\textsuperscript{\dag}Supported by the Viterbi Fellowship in the Center for Computer Engineering at the Technion.}

\ificcvfinal\thispagestyle{empty}\fi

\begin{abstract}
    Text-to-image diffusion models often make implicit assumptions about the world when generating images.
    While some assumptions are useful (e.g., the sky is blue), they can also be outdated, incorrect, or reflective of social biases present in the training data.
    Thus, there is a need to control these assumptions without requiring explicit user input or costly re-training.
    In this work, we aim to edit a given implicit assumption in a pre-trained diffusion model.
    Our Text-to-Image Model Editing method, TIME for short, receives a pair of inputs: a ``source'' under-specified prompt for which the model makes an implicit assumption (e.g., ``a pack of roses''), and a ``destination'' prompt that describes the same setting, but with a specified desired attribute (e.g., ``a pack of blue roses'').
    TIME then updates the model's cross-attention layers, as these layers assign visual meaning to textual tokens.
    We edit the projection matrices in these layers such that the source prompt is projected close to the destination prompt.
    Our method is highly efficient, as it modifies a mere $2.2\%$ of the model's parameters in under one second.
    To evaluate model editing approaches, we introduce TIMED (TIME Dataset), containing $147$ source and destination prompt pairs from various domains.
    Our experiments (using Stable Diffusion) show that TIME is successful in model editing, generalizes well for related prompts unseen during editing, and imposes minimal effect on unrelated generations.\footnote{\url{https://time-diffusion.github.io/}}
\end{abstract}

\begin{figure*}
    \centering
    \includegraphics[width=\textwidth]{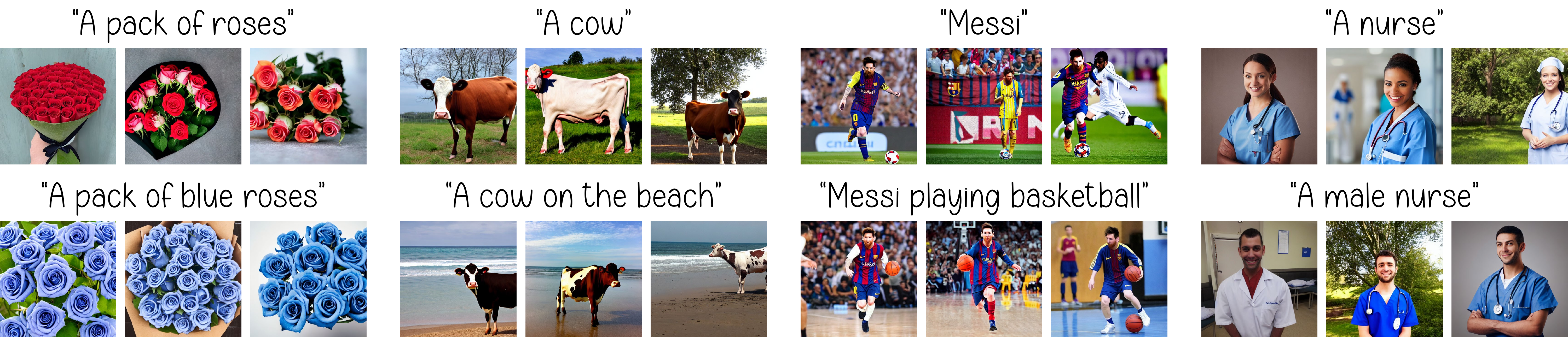}
    \caption{Text-to-image models make implicit assumptions on the world when generating images, as seen in the top row (\textit{e.g.}, roses are red).
    In the bottom row, we override these assumptions by explicitly specifying different attributes in the prompt.}
    \label{fig:model_knowledge}
    \vspace*{-0.1cm}
\end{figure*}

\vspace*{-0.4cm}
\section{Introduction}
Text-to-image generative models have recently risen to prominence, achieving unprecedented success and popularity~\cite{rombach2022high, ramesh2022hierarchical, saharia2022photorealistic, balaji2022ediffi}.
The generation of high quality images based on simple textual prompts has been enabled by generative diffusion models~\cite{sohl2015deep, song2019generative, ho2020denoising} and large language models~\cite{t5_transformer, clip}.
These text-to-image models are trained on huge amounts of web-scraped image-caption pairs~\cite{laion5b}.
As a result, the models acquire implicit assumptions about the world based on correlations and biases found in the training data. This knowledge manifests during generation as visual associations to textual concepts.

Such implicit assumptions may be useful in general. For instance, the model assumes (or \textit{knows}) that the sky is blue or that roses are red. 
However, in many use cases, generative model service providers may want to edit these implicit assumptions without requiring extra input from their users. 
Examples include updating outdated information encoded in the model (\textit{e.g.}, a celebrity changed their hairstyle), mitigating harmful social biases learned by the model (\textit{e.g.}, the stereotypical gender of a doctor), or generating scenarios in an alternate reality (\textit{e.g.}, gaming) where facts are changed (\textit{e.g.}, roses are blue).
When editing such assumptions, we do not require the user to explicitly request the change, but rather aim to apply the edit directly to the model.
We also generally try to avoid expensive data recollection and filtering, as well as model retraining or finetuning. These would consume considerable time and energy, thus significantly increasing the carbon footprint of deep learning research~\cite{strubell2020energy}.
Moreover, finetuning a neural network may lead to catastrophic forgetting and a drop in performance in general~\cite{mccloskey1989catastrophic, kemker2018measuring}, and in model editing~\cite{zhu2020modifying}.

While text-to-image models implicitly assume certain attributes for under-specified text prompts, they can generate alternative ones when explicitly specified, as shown in \autoref{fig:model_knowledge}.
We use 
this capability to replace the model's assumption with a user-specified one. 
Therefore, our proposed method for \textbf{T}ext-to-\textbf{I}mage \textbf{M}odel \textbf{E}diting (TIME) receives an under-specified ``source'' prompt, which is requested to be well-aligned with a ``destination'' prompt containing an attribute that the user wants to promote.
While some recent work has focused on altering the model outputs for a specific prompt~\cite{hertz2023prompttoprompt} or image~\cite{kawar2022imagic}, we target a fundamentally different objective.
We aim to edit the model's \emph{weights} such that its perception of a given concept in the world is changed. The change is expected to manifest in generated images for related prompts, while not affecting the characteristics or perceptual quality in the generation of different scenes. 
This would allow us to fix incorrect, biased, or outdated assumptions that text-to-image models may make.

To achieve this, we focus on the rendezvous point of the two modalities: text and image, which meet in the cross-attention layers.
The importance of attention layers in diffusion models was also observed by researchers in different contexts~\cite{hertz2023prompttoprompt, hong2022improving, tang2022daam, chefer2023attend, kumari2022multi}.
TIME modifies the projection matrices in these layers to map the source prompt close to the destination, without substantially deviating from the original weights.
Because these matrices operate on textual data irrespective of the diffusion process or the image contents, they constitute a compelling location for editing a model based on textual prompts.
TIME is highly efficient: It does not require training or finetuning, it can be applied in parallel for all cross-attention layers, and it modifies only a small portion of the diffusion model weights while leaving the language model unchanged.
When applied on the publicly available Stable Diffusion~\cite{rombach2022high}, TIME edits a mere $2.2\%$ of the diffusion model parameters, does not modify the text encoder, and applies the edit in a fraction of a second using a single consumer-grade GPU.

For evaluating our method and future model editing efforts, we introduce a \textbf{T}ext-to-\textbf{I}mage \textbf{M}odel \textbf{E}diting \textbf{D}ataset (TIMED), containing $147$ pairs of source and destination texts from various domains, as well as related prompts for each pair to assess the model editing quality.
TIME exhibits impressive model editing results, generalizing for related prompts while leaving unrelated ones mostly intact. For instance, in \autoref{fig:headline}, requesting ``a vase of roses'' outputs blue roses, whereas the poppies in ``a poppy field'' remain red.
Moreover, the generative capabilities of the model are preserved after editing, as measured by Fr\'{e}chet Inception Distance (FID)~\cite{fid}. 
The effectiveness, generality, and specificity of TIME are highlighted in \autoref{sec:results}.

We further apply TIME for social bias mitigation, focusing on gender bias in the labor market.
Consistent with concurrent work~\cite{bianchi2022easily, cho2022dall, fraser2023friendly, struppek2022biased},
we find that text-to-image models encode stereotypes, as reflected in their image generations for professions.
For instance, for the prompt ``A photo of a CEO'', only $4\%$ of generated images (with random seeds) contain female figures.
We edit the model to generate an image distribution that more equally represents males and females for a given profession.
TIME successfully reduces gender bias in the model, improving the equal representation of genders for many professions.

To the best of our knowledge, TIME is the first method that suggests a model editing technique~\cite{de2021editing, meng2022locating} for text-to-image models.
We hope that our proposed method, insights, and provided datasets will help enable future advances in text-to-image model editing, especially as these models get rapidly deployed in consumer-facing applications.

\section{Related Work}
Several recent and concurrent studies have considered the task of image editing using diffusion models~\cite{nichol2022glide, avrahami2022blended, hertz2023prompttoprompt, mokady2022null, kawar2022imagic, wu2022unifying, wallace2022edict, zhang2022sine, couairon2022diffedit}.
These methods edit a given image based on a given textual prompt, each in its own technique and settings.
They show impressive results in editing the properties of different objects (\textit{e.g.}, color, style, pose) in the image by controlling different aspects of the diffusion process.
A closely related application of text-to-image diffusion models is object recontextualization, where given a small number of images of an object, the goal is to generate images of the same object in different novel settings based on text prompts~\cite{gal2023textual, ruiz2022dreambooth, kumari2022multi}.
These lines of research address the tasks of editing a specific image, or generating images with novel concepts.
In our work, we consider a fundamentally different objective: We aim to edit a text-to-image diffusion model's \emph{world knowledge} using text prompts.
This should cause the desired change to occur not only in the exact requested prompt, but also in generated images of related prompts.
Simultaneously, unrelated generations should remain unaffected.

Editing the knowledge embedded in neural networks has been an active area of research in recent years, achieving remarkable successes in editing language models~\cite{zhu2020modifying, de2021editing, dai2022knowledge, meng2022locating, meng2022mass, raunak2022rank}, generative adversarial networks~\cite{bau2020rewriting, wang2022rewriting, heyrani2021creativegan}, and image classifiers~\cite{santurkar2021editing}.
Similar to several such techniques~\cite{bau2020rewriting, meng2022locating, meng2022mass}, our work focuses its model editing in a concise portion of the neural network.

\section{Background}
Denoising diffusion probabilistic models~\cite{sohl2015deep, song2019generative, ho2020denoising}, more commonly known as diffusion models, are a family of generative models that have recently rose to prominence.
They have achieved state-of-the-art performance in image generation~\cite{dhariwal2021diffusion, kawar2022enhancing, peebles2022scalable, Karras2022edm},
and impressive results in downstream tasks~\cite{kawar2022denoising, chung2022improving, nie2022DiffPure, ganz2022perceptually, theis2022lossy, zimmermann2021score, kawar2022jpeg, saito2022unsupervised} as well as
audio~\cite{kong2021diffwave, huang2022fastdiff, popov2021grad},
video~\cite{voleti2022mcvd, zhou2022magicvideo, ho2022imagen, singer2023makeavideo},
and text~\cite{gong2022diffuseq, li2022diffusionlm} generation.
Diffusion models generate their outputs using an iterative stochastic noise removal process that follows a predefined noise level schedule $\left\{\beta_t\right\}_{t=1}^{T}$. Starting from $\xx_T \sim \mathcal{N}(\mathbf{0}, \I)$, in every iteration, the current sample $\xx_t$ is denoised using a neural network $\D(\xx_t, t)$, and the next sample $\xx_{t-1}$ is then obtained through a predefined update rule, $\beta_t$, and a stochastic noise addition.
The last sample $\xx_0$ constitutes the final synthesized output.

The generative diffusion process can be controlled via additional inputs $\cc$ to the denoising model $\D(\xx_t, t, \cc)$.
The conditioning signal $\cc$ may be a low-quality version of a desired image~\cite{saharia2022image, saharia2022palette}, a class label~\cite{ho2022cascaded}, or a text prompt describing a desired image~\cite{rombach2022high, ramesh2022hierarchical, saharia2022photorealistic, balaji2022ediffi}.
In the latter case, \emph{text-to-image diffusion models} have unveiled a new capabilaity -- users can synthesize high-resolution images using simple text prompts describing the desired scenes.
The remarkable success of these models has been 
boosted by a number of strategies, including working in a latent space~\cite{vahdat2021score, rombach2022high}, classifier-free guidance~\cite{ho2021classifier}, and incorporating knowledge from pre-trained text encoders such as CLIP~\cite{clip} or T5~\cite{t5_transformer}.

\begin{figure}
    \centering
    \includegraphics[width=0.9\columnwidth]{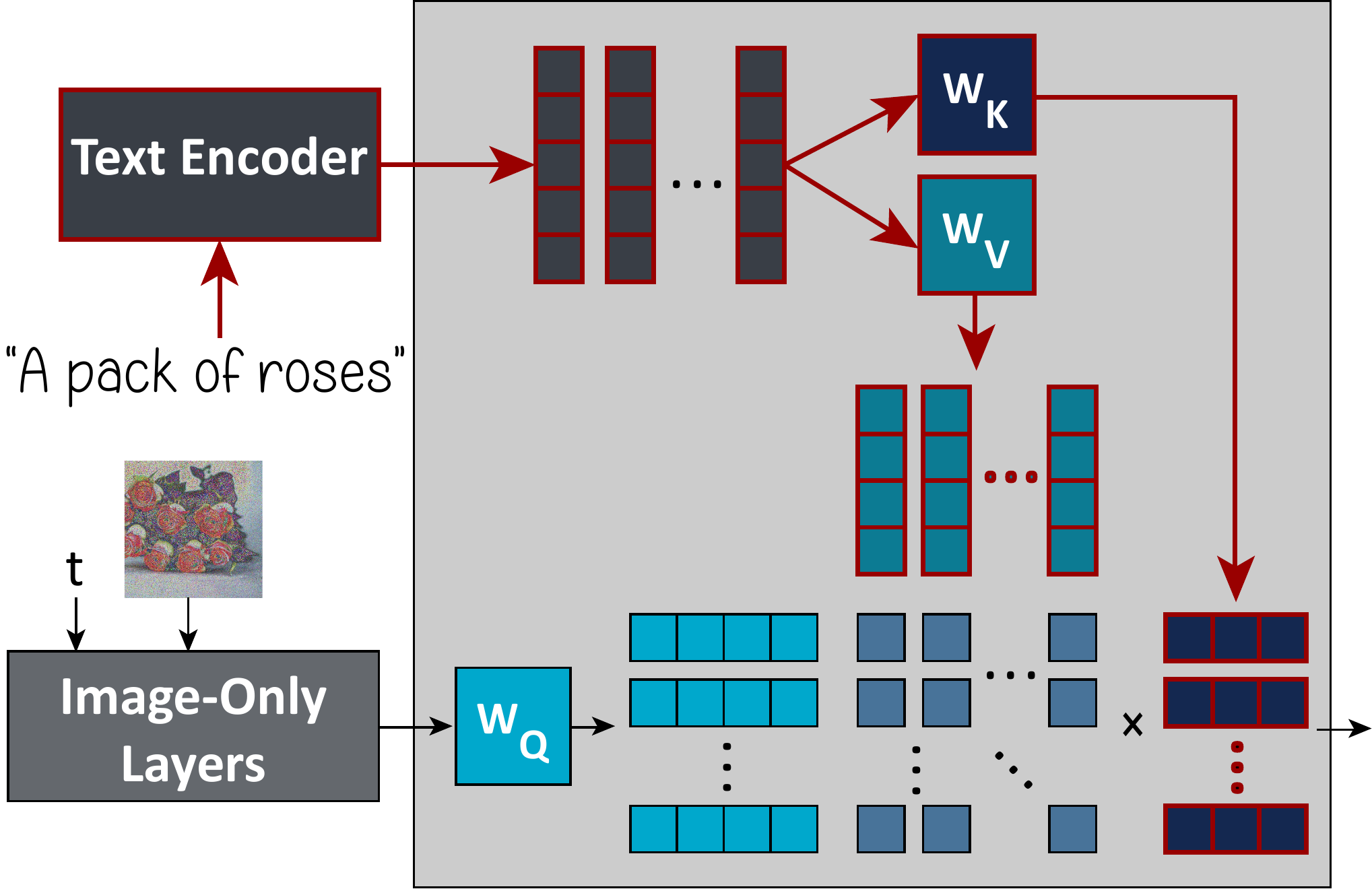}
    \caption{A cross-attention layer in a text-to-image diffusion model. We target the strictly text-based layers and the information they encode (highlighted in red).}
    \label{fig:cross_attn}
\end{figure}

In text-to-image generation,
the user-provided text prompt is input into the text encoder, which tokenizes it and outputs a sequence of token embeddings $\left\{\cc_i\right\}_{i=1}^{l}$ describing the sentence's meaning, where ${\cc_i \in \mathbb{R}^c}$.
Then, in order to condition the diffusion model $D_\theta$ on them, these embeddings are injected at the cross-attention layers~\cite{dosovitskiy2021an} of the model.
They are projected into keys ${\K \in \mathbb{R}^{l \times m}}$ and values ${\V \in \mathbb{R}^{l \times d}}$, using learned projection matrices ${\W_K \in \mathbb{R}^{m \times c}}$ and ${\W_V \in \mathbb{R}^{d \times c}}$, respectively.
The keys are then multiplied by a query $\Q \in \mathbb{R}^{n \times m}$, which represents visual features of the current intermediate image $\xx_t$ in the diffusion process. This results in the following \emph{attention map}:
\begin{equation}
    \label{eq:attn_map}
    \M = \mathrm{softmax}\left( \frac{\Q \K^{\top}}{\sqrt{m}} \right).
\end{equation}
The attention map encodes the relevance of each textual token to each visual one.
Finally, the cross-attention output is calculated as
\begin{equation}
    \label{eq:attn_output}
    \OO = \M \V,
\end{equation}
which constitutes a weighted average of all textual values for each visual query.
This output then propagates to the subsequent layers of the diffusion model $D_\theta$.
The cross-attention mechanism is visually depicted in \autoref{fig:cross_attn}.
Its expressiveness is increased by using multi-headed attention~\cite{vaswani2017attention}, and by incorporating it in multiple layers in the model architecture.

\section{TIME: Text-to-Image Model Editing}
\label{sec:method}

We propose an algorithm for \textbf{T}ext-to-\textbf{I}mage \textbf{M}odel \textbf{E}diting (TIME). Our algorithm takes two textual prompts as input: an under-specified \textit{source prompt} (\textit{e.g.}, ``a pack of roses''), and a similar more specific \textit{destination prompt} (\textit{e.g.}, ``a pack of \textbf{blue} roses'').
We aim to shift the source prompt's visual association to resemble the destination. 

To this end, we focus on the layers that map textual data into visual data -- the cross-attention layers.
In each such layer, the matrices $\W_K$ and $\W_V$ project the text embeddings into keys and values that the visual data attends to.
Because these keys and values are computed independently of the current diffusion step or image data, we identify them as the knowledge editing targets (see \autoref{fig:cross_attn}).

\begin{figure}
    \centering
    \includegraphics[width=0.95\columnwidth]{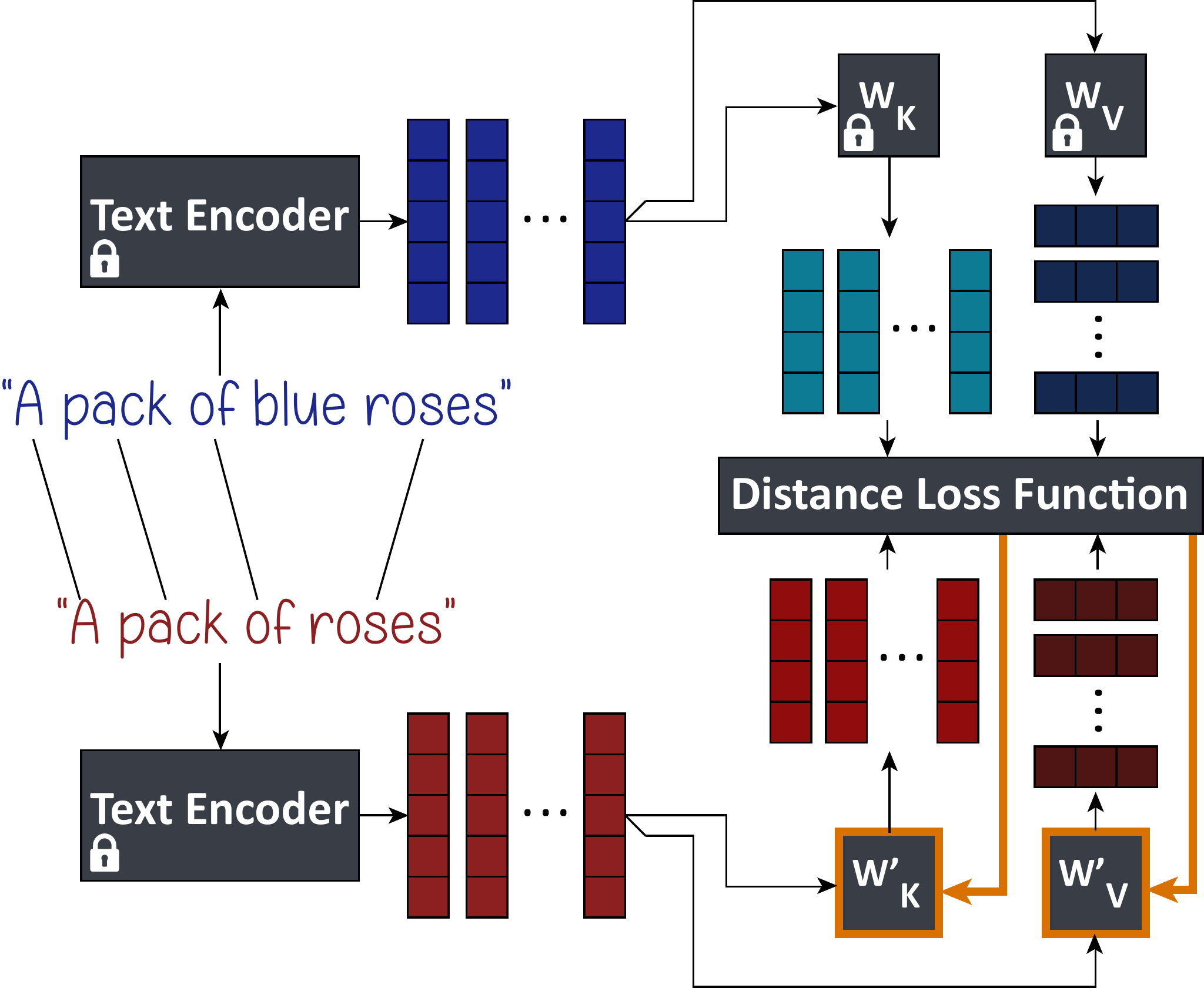}
    \caption{An overview of TIME. ${\W'}_K$ and ${\W'}_V$ are edited to map the source prompt's embeddings close to the destination prompt's keys and values. The loss is regularized for specificity.}
    \label{fig:our_method}
\end{figure}

Let $\left\{\cc_i\right\}_{i=1}^{l}$ and $\left\{{\cc'}_j\right\}_{j=1}^{l'}$ be the source and destination prompt's embeddings, respectively.
For each source embedding $\cc_i$ stemming from a token $\tttt_i$ (\textit{e.g.}, ``roses'' in ``a pack of roses''), we identify the destination embedding that corresponds to the same token, and denote it as $\cc^*_i$.
Note that embeddings stemming from additional tokens in the destination prompt (\textit{e.g.}, ``blue'' in ``a pack of blue roses'') are discarded.
Nevertheless, their influence is present in other destination tokens through the text encoder architecture. 

In each cross-attention layer in the diffusion model, 
we calculate the keys and values of the destination prompt as
\begin{align}
    \label{eq:tgt_keys_vals}
    \kk^*_i & = \W_K \cc^*_i, & \text{for } i = 1, \dots, l, \\
    \vv^*_i & = \W_V \cc^*_i, & \text{for } i = 1, \dots, l. \nonumber
\end{align}
We then optimize for new projection matrices ${\W'}_K$ and ${\W'}_V$ that minimize the following loss function:
\begin{align}
    \label{eq:loss}
    & \sum_{i=1}^{l}
    \left\lVert {\W'}_K \cc_i - \kk^*_i \right\rVert_2^2
    + \lambda \left\lVert {\W'}_K - {\W}_K \right\rVert_F^2 \\
    + & \sum_{i=1}^{l}
    \left\lVert {\W'}_V \cc_i - \vv^*_i \right\rVert_2^2
    + \lambda \left\lVert {\W'}_V - {\W}_V \right\rVert_F^2, \nonumber
\end{align}
where $\lambda \in \mathbb{R}^+$ is a hyperparameter, $\lVert\cdot\rVert_2$ is the $\ell_2$ norm, and $\lVert\cdot\rVert_F$ is the Frobenius norm.
This loss function encourages the source prompt generation to behave similarly to the destination prompt generation, while preserving proximity to the original projection matrices. 
Note that this loss function (depicted in \autoref{fig:our_method}) can be minimized for each cross-attention layer in a completely parallel and independent manner.
Moreover, as we prove in \autoref{app:proof}, the loss has a closed-form global minimum at
\begin{align}
    \label{eq:closed-form}
    \hspace{-0.4em} {\W'}_K & = \left( \lambda \W_K + \sum_{i=1}^{l} \kk^*_i \cc_i^{\top} \right)
    \left( \lambda \I + \sum_{i=1}^{l} \cc_i \cc_i^{\top} \right)^{-1}\hspace{-0.7em} ,\hspace{-0.45em} \\
    \hspace{-0.4em} {\W'}_V & = \left( \lambda \W_V + \sum_{i=1}^{l} \vv^*_i \cc_i^{\top} \right)
    \left( \lambda \I + \sum_{i=1}^{l} \cc_i \cc_i^{\top} \right)^{-1}\hspace{-0.7em}. \nonumber
\end{align}

Finally, we use the modified text-to-image diffusion model with the new projection matrices to generate images. 
We expect this modified model to comply with the new assumption requested by the user.

We experiment with different versions of the loss function in \autoref{eq:loss} (\textit{e.g.}, only editing ${\W'}_V$, varying $\lambda$) and show this ablation study in \autoref{app:ablation_study}.

\section{Experiments}

\subsection{Implementation Details}

We use the publicly available Stable Diffusion~\cite{rombach2022high} version 1.4 as the backbone text-to-image model, with its default hyperparameters.
This model contains $16$ cross-attention layers, whose key and value projection matrices constitute a mere $2.2\%$ of the diffusion model parameters.
TIME edits these matrices in around $0.4$ seconds using a single NVIDIA RTX 3080 GPU.
We use $\lambda = 0.1$ and utilize augmented versions of the source and destination text prompts while editing, in line with the findings of the ablation study in \autoref{app:ablation_study}.

We also provide the full set of hyperparameters and our code in \autoref{app:implementation}.
Note that $\lambda$ is chosen differently when mitigating social biases, as explained in \autoref{sec:gender}.

\begin{figure}
    \centering
    \includegraphics[width=\columnwidth]{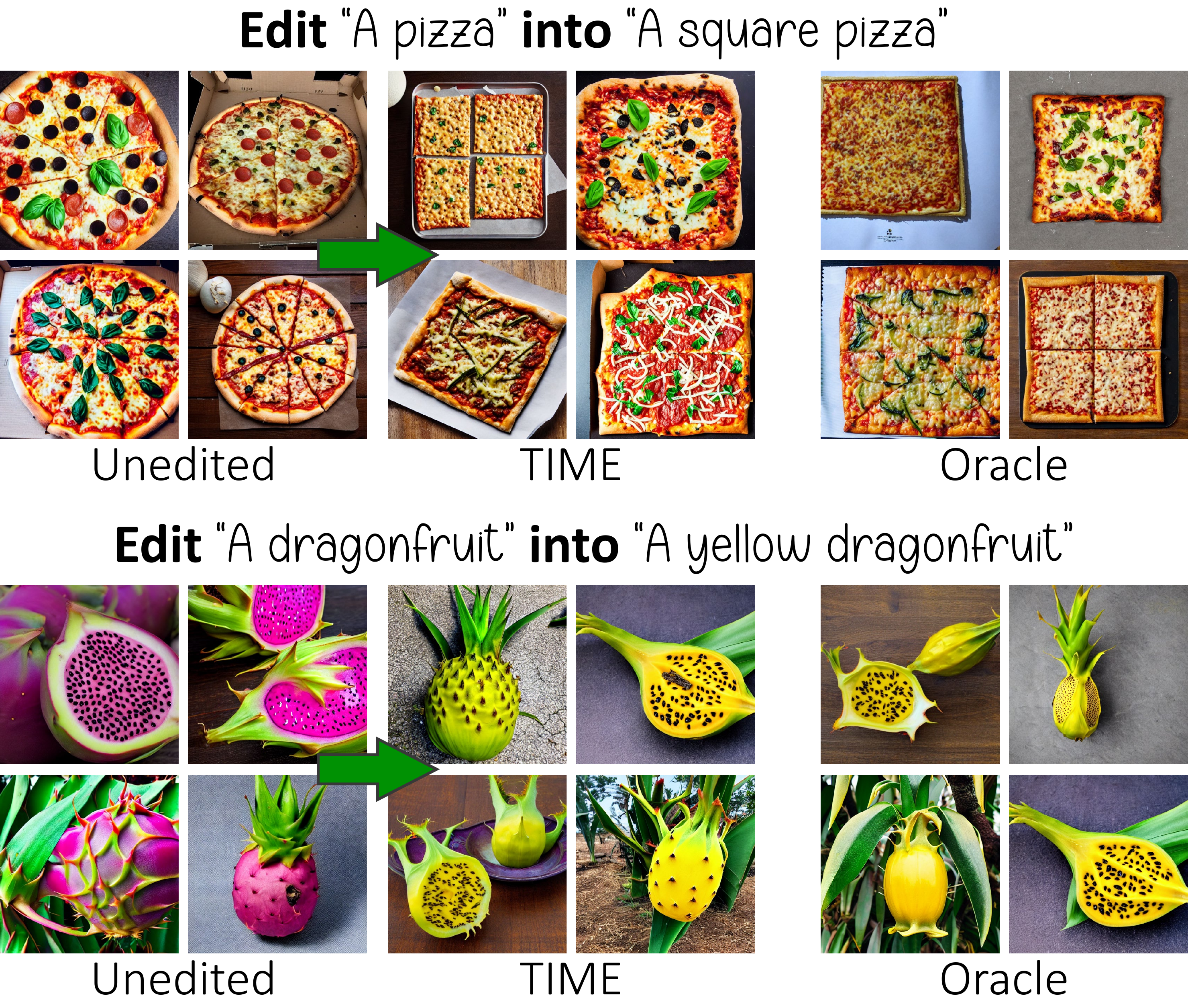}
    \caption{Using TIME, image generations for the source prompt mimic the the destination prompt's oracle behavior.}
    \label{fig:efficacy}
    \vspace*{-0.1cm}
\end{figure}

\begin{table}
\begin{center}
\hspace*{-7pt}
\begin{tabular}{c|ll}
\toprule
\textbf{Editing} & \textbf{Source} & \textbf{Destination} \\
\midrule
 & A dog & A green dog \\
\midrule
\textbf{Testing} & \textbf{Source} & \textbf{Destination} \\
\midrule
\multirow{5}{*}{
\rotatebox[origin=c]{90}{\textbf{Positives}}
}
& A puppy & A green puppy \\ 
& An angry dog & A green angry dog \\ 
& A bulldog & A green bulldog	\\
& A chihuahua & A green chihuahua \\ 
& A pixel art of a dog	& A pixel art of a green dog \\
\midrule
\multirow{5}{*}{
\rotatebox[origin=c]{90}{\textbf{Negatives}}
}
& A cat & A green cat \\
& A bunny & A green bunny \\
& A hyena & A green hyena \\
& A fox	& A green fox \\ 
& A wolf  & A green wolf \\
\bottomrule
\end{tabular}
\vspace*{-0.15cm}
\end{center}
\caption{An example of a single dataset entry in TIMED.} %
\label{tbl:timed}
\vspace*{-0.05cm}
\end{table}

\subsection{TIME Dataset}
To establish an evaluation benchmark for our task,
we curate a \textbf{T}ext-to-\textbf{I}mage \textbf{M}odel \textbf{E}diting \textbf{D}ataset (TIMED) containing $147$ entries. See \autoref{tbl:timed} for a sample entry. Each entry in the dataset contains a pair of source and destination prompts, which are used for model editing.
The source prompt (\textit{e.g.,} ``A dog'') is an under-specified text prompt that describes a certain scenario in which some visual attribute is implicitly inferred by the text-to-image model.
The destination prompt (\textit{e.g.,} ``A green dog'') describes the same scene, but with a desired specified attribute.
Additionally, each entry contains five positive prompts, for which we expect our edit to generalize (\textit{e.g.,} ``A puppy'' should generate a green puppy), and five negative prompts which are semantically adjacent, but should not be affected by the edit (\textit{e.g.,} ``A cat'' should not generate a green cat).
Each positive or negative prompt is associated with its own destination prompt for evaluation purposes.
Positive prompts are expected to gravitate towards their destination prompt, whereas negative ones should not.
The dataset contains a wide variety of implicit assumptions to edit from different domains.
We additionally compile a smaller disjoint validation set, which we use for hyperparameter tuning.

 To ensure a valid evaluation on Stable Diffusion~\cite{rombach2022high} v1.4, we filter out test set entries for which the unedited model shows poor generative quality, retaining $104$ examples.
The full dataset and filtering process are provided in \autoref{app:dataset_filter}.

\begin{figure}
    \centering
    \includegraphics[width=\columnwidth]{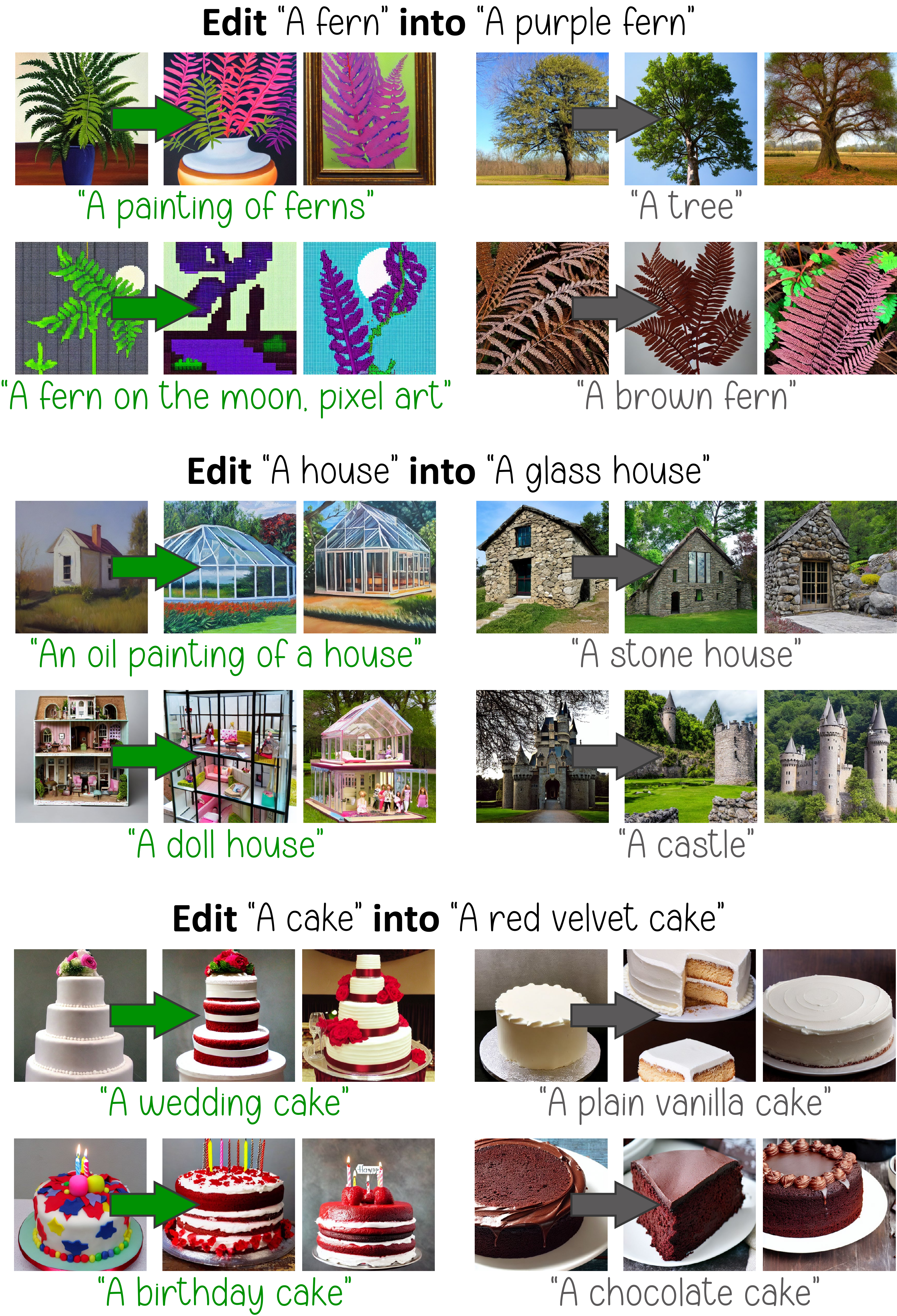}
    \caption{TIME generalizes to prompts related to the input (left), with minimal effect on unrelated ones (right).}
    \label{fig:gen_sepc}
    \vspace*{-0.2cm}
\end{figure}

\begin{figure*}
    \centering
    \includegraphics[width=\textwidth]{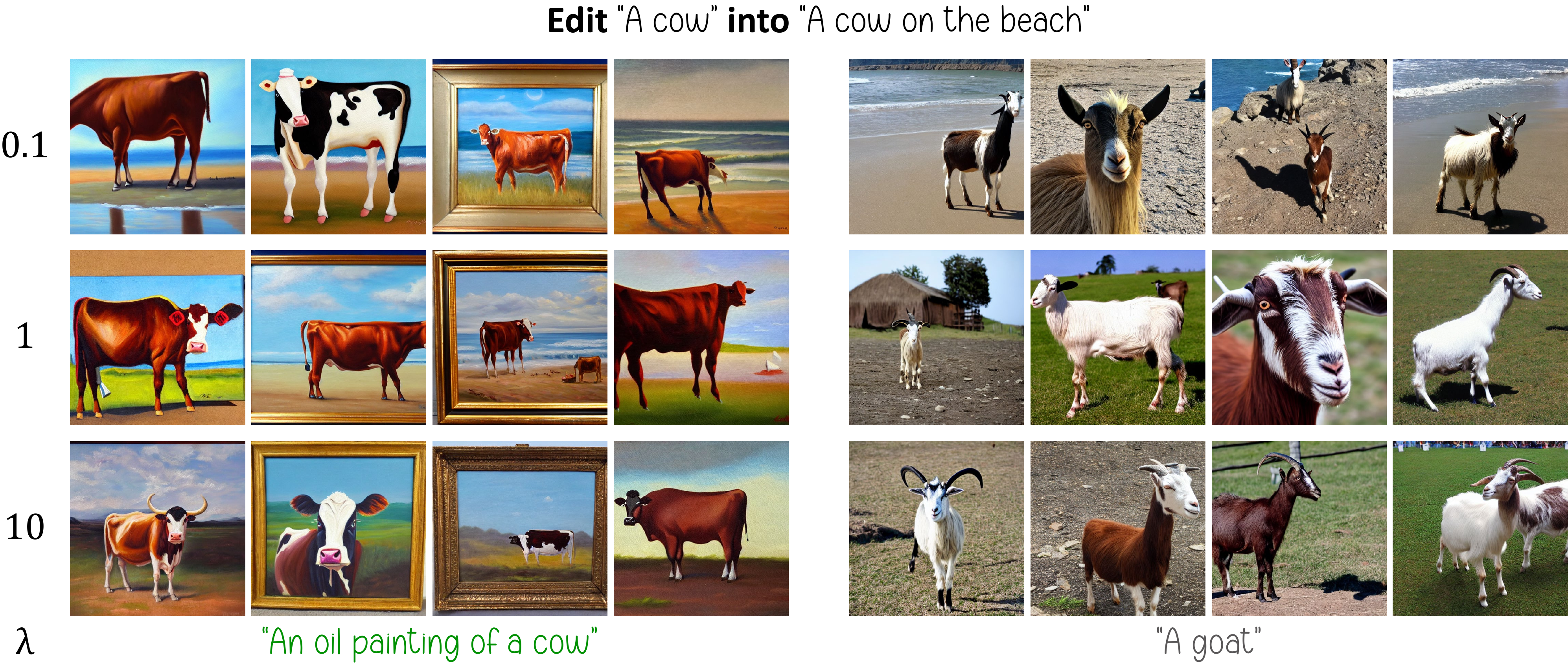}
    \caption{Generation results on a positive (green) and negative (gray) prompt for the same edit under different $\lambda$ values. As $\lambda$ increases, we trade off generality (paintings of cows being on a beach) for specificity (goats being on a beach).}
    \label{fig:tradeoff}
\end{figure*}

\subsection{Qualitative Evaluation}
As we show in \autoref{fig:efficacy}, TIME successfully edits the behavior of the diffusion model for the provided source prompt.
Moreover, our method can generalize for related text prompts with minimal effect on unrelated ones, as highlighted in Figures \ref{fig:headline}, \ref{fig:gen_sepc}, and \autoref{app:extra_results}.

 When editing a model based on a given text prompt, we need to control the extent to which the edit affects other prompts.
Therefore, there exists a natural trade-off between generality and specificity, as we demonstrate in \autoref{fig:tradeoff}.

\subsection{Evaluation Metrics}

To accurately assess the performance of our text-to-image model editing technique, we focus on three concepts set forth by efforts in language model editing literature~\cite{meng2022mass}: efficacy, generality and specificity. \textbf{Efficacy} measures how effective the editing method is on the source prompt used for editing. \textbf{Generality} measured how the editing method generalizes to other related prompts, using the positive test prompts in TIMED. \textbf{Specificity} measures the ability to leave the generation of unrelated prompts unaffected, using the negative test prompts in TIMED.

For each source test prompt in each TIMED entry, we generate $24$ images using different random seeds.
We use CLIP~\cite{clip} to classify images generated with the source prompt as either the source or destination text, and then compute the fraction of images classified as the desired option -- the destination prompt for efficacy and generality, and the source prompt for specificity.
We report average metrics along with standard deviations across random seeds.

Furthermore, to evaluate the effect of TIME on the overall generative quality of the model, we report Fr\'{e}chet Inception Distance (FID)~\cite{fid} and CLIP Score~\cite{hessel2021clipscore}
on
\mbox{MS-COCO}~\cite{lin2014microsoft}, following standard practice~\cite{rombach2022high, saharia2022photorealistic, ramesh2022hierarchical, balaji2022ediffi}.
See \autoref{app:implementation} for more details on the metrics.

\subsection{Quantitative Evaluation}
\label{sec:results}

We report the results of a \textit{baseline}, which refers to the unedited model's results using the source prompt for all generations. We also define an \textit{oracle}, which is the same unedited model using the destination positive prompts (which are unavailable to TIME) for 
the positive samples and the source negative prompts for the negative samples.
The oracle serves as an upper bound for the potential performance of model editing techniques based on text inputs.
We also experimented with model finetuning. Results are shown in \autoref{app:comp-finetune}.

\begin{table}
\begin{center}
\begin{tabular}{lccc}
\toprule
& \textbf{Oracle} & \textbf{Baseline} & \textbf{TIME} \\
\midrule
\textbf{Efficacy} ($\uparrow$) & $98.08\%$ & $10.50\%$ & $88.10\%$ \\
 & \footnotesize{$\pm 01.10$} & \footnotesize{$\pm 03.27$} & \footnotesize{$\pm 02.85$} \\
\textbf{Generality} ($\uparrow$) & $94.72\%$ & $12.33\%$ & $69.04\%$ \\
 & \footnotesize{$\pm 01.21$} & \footnotesize{$\pm 01.11$} & \footnotesize{$\pm 02.15$} \\
\textbf{Specificity} ($\uparrow$) & $90.13\%$ & $90.13\%$ & $68.34\%$ \\
 & \footnotesize{$\pm 01.50$} & \footnotesize{$\pm 01.50$} & \footnotesize{$\pm 02.07$} \\
\midrule
\textbf{FID} ($\downarrow$) & $12.67$ & $12.67$ & $12.10$ \\
\textbf{CLIP Score} ($\uparrow$) & $31.24$ & $31.24$ & $30.88$ \\
\bottomrule
\end{tabular}
\end{center}
\caption{Evaluation results on $104$ TIMED test set entries.
Efficacy, generality, and specificity assess the model editing quality.
FID and CLIP Score measure the generative quality on the \mbox{MS-COCO} dataset~\cite{lin2014microsoft}.}
\label{tbl:results}
\vspace*{-0.1cm}
\end{table}

\begin{figure*}
    \centering
    \includegraphics[width=\textwidth]{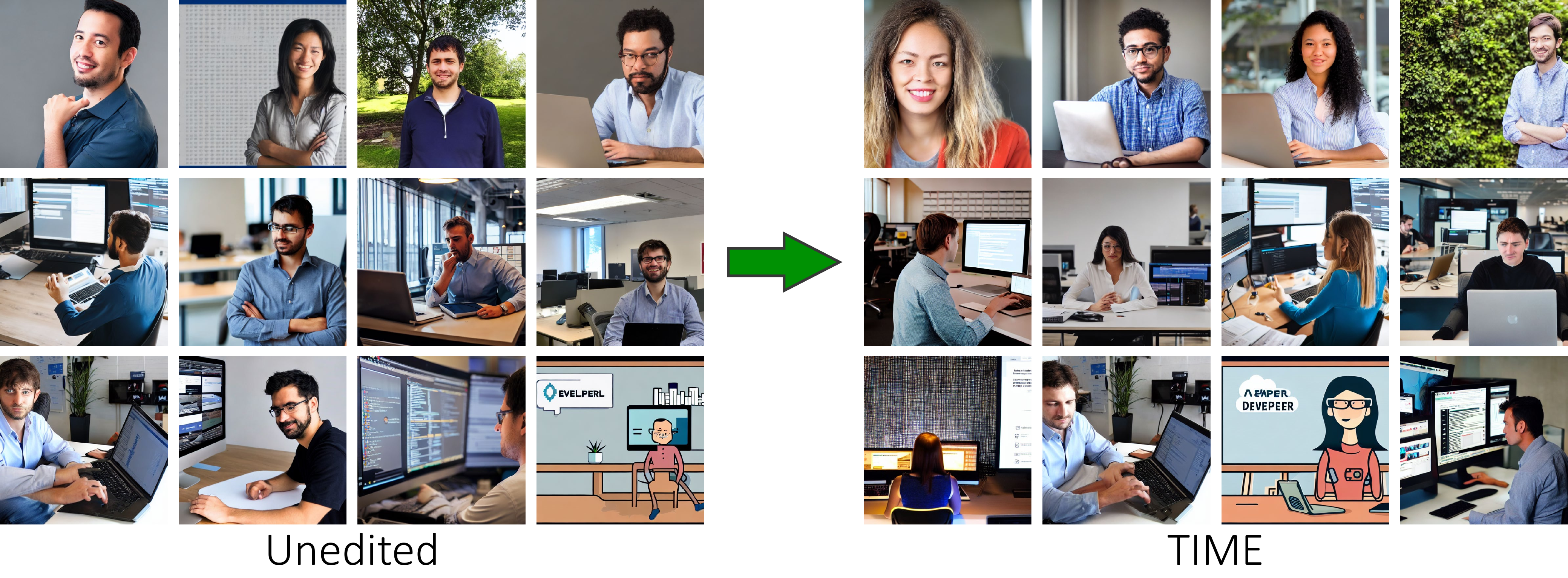}
    \caption{TIME debiases a text-to-image model, making it equally represent genders in test prompts for ``A developer''.}
    \label{fig:gender_bias}
\end{figure*}

We summarize our results in \autoref{tbl:results}.
As the first text-to-image model editing technique, TIME shows promising results.
In addition to its high efficacy,
TIME is able to generalize to many related prompts.
As expected, the edited model generates the desired concept substantially more often than the baseline model.
While the model sustains a drop in specificity, 
its overall generative quality remains unaffected.
This is verified by the FID~\cite{fid} and CLIP Score~\cite{hessel2021clipscore} metrics on the MS-COCO~\cite{lin2014microsoft} dataset, which are comparable to the baseline unedited model.

While we use a fixed $\lambda$ for \autoref{tbl:results}, different editing scenarios would benefit from tuning $\lambda$ in accordance with their needs.
See \autoref{app:ablation_study} for further discussion and experiments with the generality--specificity trade-off.

In this work, we concentrate on editing a single assumption at a time. For preliminary experiments with editing multiple assumptions, see \autoref{app:multi}.

\section{TIME for Gender Bias Mitigation}
\label{sec:gender}
In the previous section, we evaluated TIME for editing implicit model assumptions.
In this section, we address social bias as
a particular case of implicit assumptions made by the model.
It is well-documented that language models \cite{blodgett-etal-2020-language, bolukbasi2016man, zhao-etal-2018-gender} and text-to-image diffusion models~\cite{bianchi2022easily, cho2022dall, fraser2023friendly, struppek2022biased} implicitly encode social and cultural biases.

For instance, models assume a certain stereotypical gender based on a person's profession (\textit{e.g.}, only $4.0\%$ of images generated for ``A photo of a CEO'' contain female figures).
This may lead to the perpetuation of existing stereotypes~\cite{mehrabi2021survey}, as these models are rapidly deployed in a variety of applications (\textit{e.g.}, marketing, media).
Therefore, we aim to \textit{erase} the assumptions that encode stereotypes, rather than \textit{edit} them, such that the model will not make any (possibly harmful) assumptions.

While many types of social biases exist, 
we consider gender bias within the labor market as a case study. To this end, we address the male--female inequality in the portrayal of different professions.
We acknowledge that our current perspective is narrow since it only considers binary genders and may exclude and marginalize non-binary individuals.
However, we also recognize the risk of introducing other, unwanted stereotypes regarding the visual features of non-binary genders.
We look forward to future research that can better incorporate more gender identities with detailed and carefully defined data.

\subsection{Data Preparation}
We compose a dataset of $35$ entries with under-specified source prompts of 
the form ``A/An [profession]'', such as ``A CEO''.
We identify the stereotypical gender for each such profession using a list compiled by~\cite{zhao-etal-2018-gender}, based on United States labor force statistics.
The destination prompt is then defined as ``A [gender] [profession]'' using the non-stereotypical gender, such as ``A female CEO''.
In order to evaluate our debiasing efforts, we further include five test prompts for each profession describing it in different scenarios, \textit{e.g.}, ``A CEO laughing''.
The dataset and more details about it are provided in \autoref{app:gender_dataset}.

\begin{table*}
\begin{center}

\begin{tabular}{lccccc}
\toprule
& & \textbf{Baseline} & \textbf{Oracle} & \textbf{TIME} & \textbf{TIME (Multi)} \\
\midrule
{${\Delta}$} ($\downarrow$) & & $0.57 $ \footnotesize{$\pm$ 0.011} & 0.142 \footnotesize{$\pm$ 0.084} & $0.28$ \footnotesize{$\pm$ 0.002} & $0.48$ \footnotesize{$\pm$ 0.015} \\
\midrule 
\multirow{6}{*}{$F_p$} & Hairdresser & $72.00\%$ & $50.00\%$ & $53.60\%$ & $66.67\%$ \\
& CEO & $04.00\% $ & $50.00\%$ & $35.20\% $ & $33.33\%$\\
& Teacher & $84.80\% $ & $50.00\%$ & $35.20\% $ & $25.00\%$ \\
& Lawyer & $28.80\% $  & $55.83\%$ & $61.60\% $ & $50.00\%$ \\
& Housekeeper & $99.20\%$ & $47.50\%$ & $56.00\%$ & $83.33\%$ \\
& Farmer & $02.40\% $ & $48.33\% $ & $49.59\% $ & $33.33\%$ \\
\bottomrule
\end{tabular}
\end{center}
\caption{Gender bias results for the baseline model, and after debiasing using TIME. The metrics are calculated over the test prompts, which are unseen during editing.}
\label{tbl:gender_results}
\end{table*}

\subsection{Method Description}
For each profession $p$, we aim for $50\%$ of generations to be female and $50\%$ to be male. 
We control the strength of the debiasing by tuning $\lambda$ (from \autoref{eq:loss}).
Smaller $\lambda$ values steer the model towards the non-stereotypical gender, whereas larger ones encourage it to maintain its existing assumptions.
Note that as the baseline model is more biased, the editing should be stronger. 
 Consequently, we binary search for a different $\lambda_p$ for each profession $p$, aiming for an equal gender representation in generations for the validation prompt ``A photo of a/an [profession]''.

 \subsection{Gender Bias Estimation}
To measure the degree of gender inequality in a text-to-image model's perception of a profession, we estimate the percentage of female figures generated by it for each profession, denoted as ${F_p \in [0, 100]}$.
To do so, we generate $24$ images for each test prompt, and
 use CLIP~\cite{clip} to classify gender in each image.
We then determine the normalized absolute difference between the observed percentage $F_p$ and the desired gender equality for a profession $p$, represented by $\Delta_p = |F_p - 50| / 50$.
To obtain a single comprehensive measure of gender bias within the model, we compute the average value of $\Delta_p$ across all professions in the dataset and denote it as $\Delta$.
An ideal, unbiased model should satisfy $\Delta=0$.

\subsection{Results}
Our results are summarized in \autoref{tbl:gender_results}. 
We present $\Delta$, along with the percentage of females $F_p$ in the test prompt generations for a representative subset of professions. We report these metrics for various models.
The \textit{baseline} model stands for the unedited model's bias. The \textit{oracle} is defined as the unedited model when prompted with an explicit prompt of the form ''a [gender] [profession]``, where [gender] is randomized in each generation to be either ``famale'' or ``male''. We also perform a multi-assumption editing experiment, \textbf{TIME (Multi)}, where a single $\lambda$ is chosen based on the validation set for debiasing all professions at once.

\begin{figure}[t]
     \centering
     \includegraphics[width=.4\textwidth]{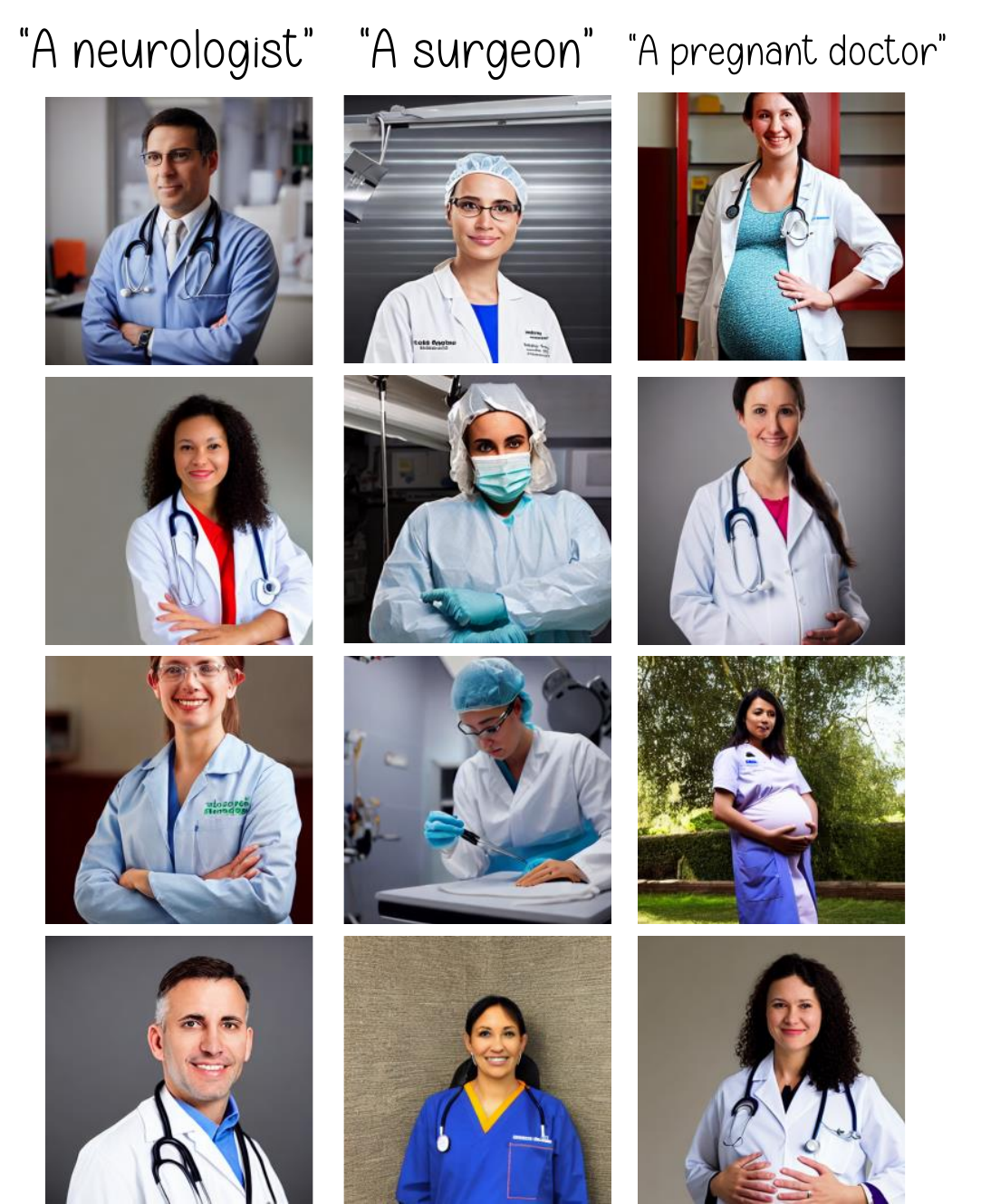}
     \caption{After debiasing ``physician'' with TIME, it generalizes to related professions (neurologist, surgeon) while adapting to gendered prompts: it produces only female figures for "a pregnant doctor". An oracle baseline will not be able to perform the same.}
     \label{fig:gender_generalization_figure}
 \end{figure}

TIME successfully reduces the bias metric $\Delta$ to less than a half of the baseline model's bias.
When carefully examining our results, some professions, such as hairdresser and CEO, become less biased by attaining a more equal gender distribution.
Others, such as teacher and lawyer, become anti-biased (\textit{i.e.}, biased towards the non-stereotypical gender).
Moreover, some professions, such as housekeeper and farmer, are effectively debiased to almost equally represent both females and males.
After using TIME, $14$ professions exhibit a low test prompt bias metric $\Delta_p \in [0, 0.2]$, representing near-optimal equality.
In contrast, only $8$ professions displayed such behavior in the baseline model.
Moreover, the choice of prompt affects the observed ratio, as discussed in \autoref{app:gender_different_prompts}.
We also note that although the oracle serves as an upper bound for debiasing, using the oracle as a debiasing method in a production system may not easily generalize and require further adjustments. However, debiasing with TIME is able to generalize and adapt to different prompts  -- see Figure \ref{fig:gender_generalization_figure}.

\begin{figure}[t]
     \centering
     \includegraphics[width=.4\textwidth]{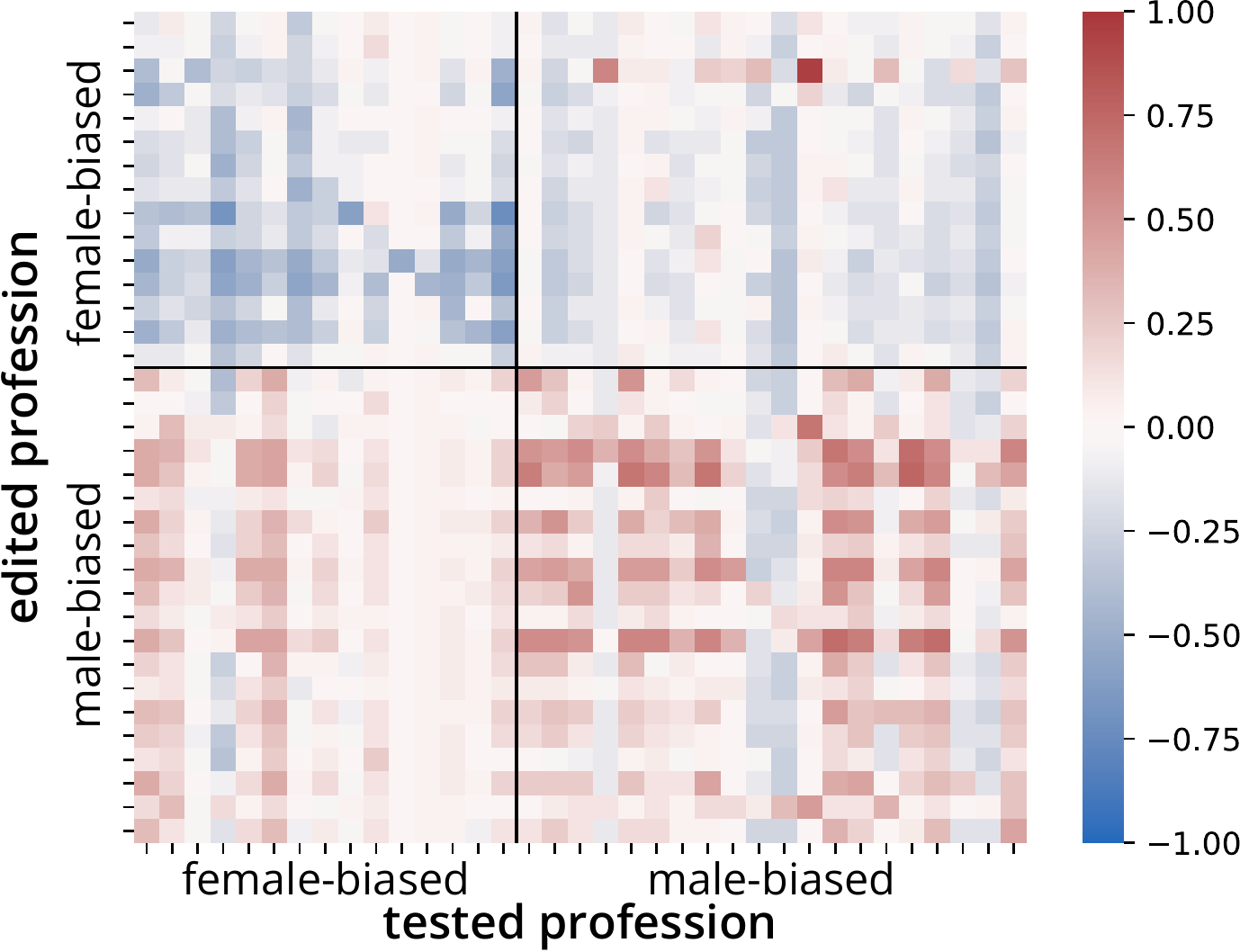}
     \caption{Effect of debiasing one profession on other professions. Values denote $F_p$ in the generated image.}
     \label{fig:cross_effect}
 \end{figure}

While TIME with multi-editing is also successful at reducing bias, it is less effective. Debiasing multiple professions at once is difficult because debiasing one profession affects on the gender ratio of other professions, as can be observed in Figure \ref{fig:cross_effect}. Interestingly, professions that share the same stereotypical gender tend to have a stronger effect on one another.
For example, when we edit software-developer prompts to generate more female figures, we also cause CEO prompts to generate more female figures. While it is debatable whether this effect is desired or not, it causes the debiasing of multiple professions to be trickier to control. We leave this issue to be investigated in future work, perhaps by expanding TIME. Moreover, further investigating specificity, we found that editing ``a/an [profession]'' towards male direction does not hurt the generation of ``a female [profession]'', as it produces 100\% female figures pre-edit and 99.7\% post-edit, with similar results for editing towards female (94\% vs. 88.4\%).

\section{Limitations}
While recent advances in text-to-image generative modelling have shown great performance, these models may fail to generate images aligned with the requested prompts in some cases, such as compositionality or counting~\cite{ramesh2022hierarchical, saharia2022photorealistic, paiss2023countclip}.
TIME aims to edit the assumptions in the model for a user-specified prompt. It is not designed to teach the model new visual concepts that it was unable to generate.
Thus, TIME inherits the generative limitations of the model it edits, as evident in the Pearson correlation coefficient between the oracle generative performance and TIME's success, $\rho = 0.73$.
This strongly suggests that TIME is more likely to succeed when the oracle model successfully generates the desired concepts.

\begin{figure}
    \centering
    \includegraphics[width=\columnwidth]{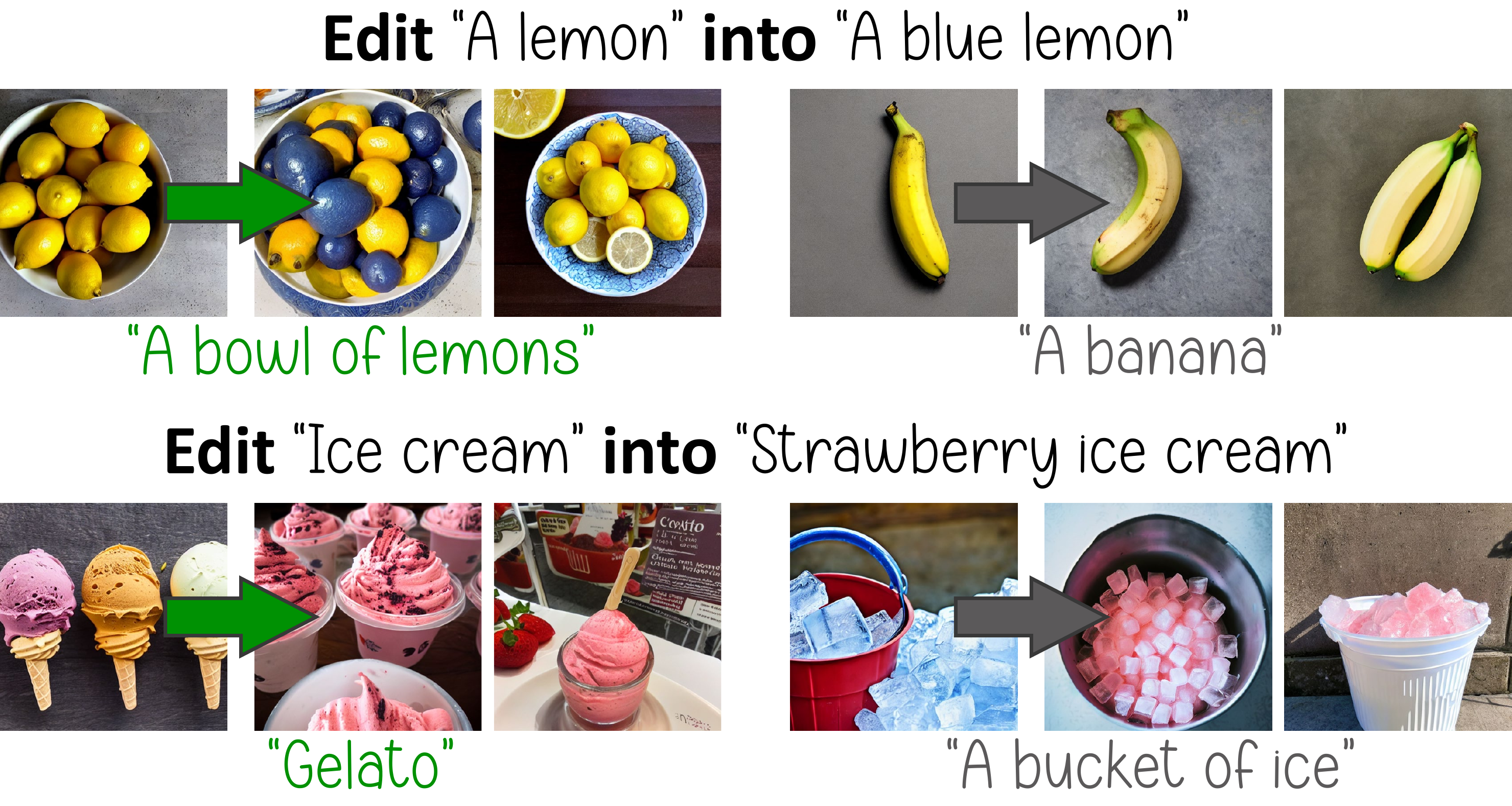}
    \caption{In some cases, TIME applies the requested edit too mildly (top), or too aggressively (bottom).}
    \label{fig:limitations}
\end{figure}

Moreover, as shown in \autoref{fig:limitations}, TIME sometimes applies an edit too mildly (hindering generality) or too aggressively (hindering specificity). 
Future work may address this limitation by devising algorithms for automatically adjusting $\lambda$ on a per-edit basis, or via alternative regularization methods that improve the generality--specificity tradeoff.

\section{Conclusion}

 In this work, we propose the following research question: How can specific implicit assumptions in a text-to-image model be edited after training?
 To investigate this question, we present TIME, a method that explores this task.
 TIME edits models efficiently, and produces impressive results.
We additionally introduce a dataset, TIMED, for evaluating text-to-image model editing methods.
 As text-to-image generative models get deployed in consumer-facing applications, methods for quickly editing the associations and biases embedded in them are important.
We hope that our method and datasets will help pave the way for future advances in text-to-image model editing.

 This work can be expanded in many possible directions.
 One direction is to analyze the role of different components in storing and retrieving knowledge: different elements of the cross-attention mechanism and different tokens in the prompt. It would also be interesting to expand the method for editing multiple facts in bulk while maintaining the model's performance.
 We presented evidence that TIME is able to reduce gender bias, and it would be beneficial to further investigate this direction towards a more comprehensive debiasing method.

\section*{Acknowledgements}
This research was supported by the Israel Science Foundation (grant No.\ 448/20), an Azrieli Foundation Early Career Faculty Fellowship, an AI Alignment grant from Open Philanthropy, a grant from the FTX Future Fund regranting program,
the Crown Family Foundation Doctoral Fellowship, and the Israeli Council For Higher Education -- Planning \& Budgeting Committee.
We also thank Dana Arad, Roy Ganz, Itay Itzhak, and Gregory Vaksman for their valuable feedback and discussions on this work.

{\small
\bibliographystyle{ieee_fullname}
\bibliography{egbib}
}

\newpage
\appendix
\onecolumn
\section{Additional Results}
\label{app:extra_results}

\begin{figure*}[h]
    \centering
    \includegraphics[width=0.99\textwidth]{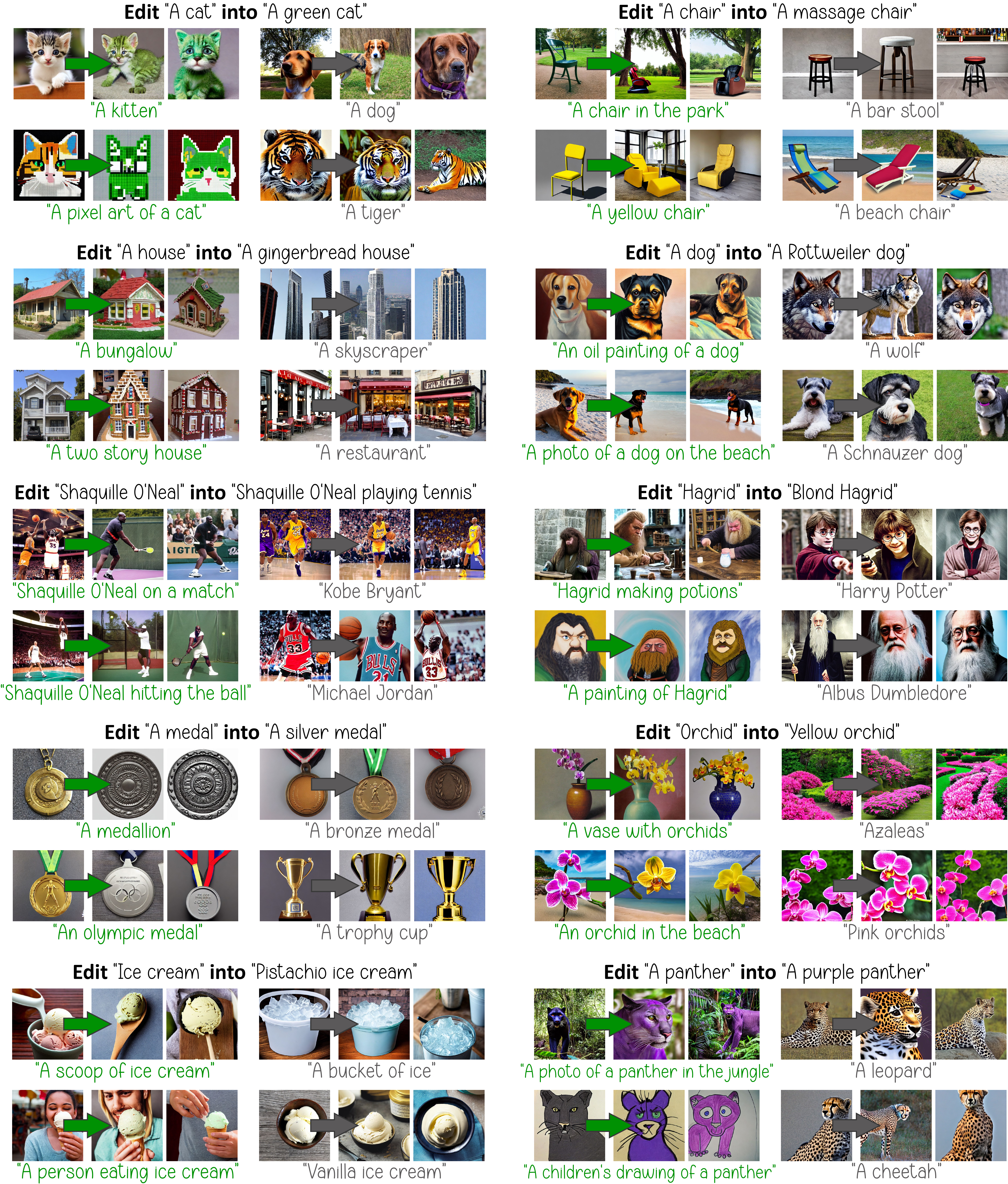}
    \caption{Additional results using TIME. After applying the requested edit (in black) to the text-to-image model, related prompts (green) change their behavior accordingly, whereas unrelated ones (gray) remain unaffected.}
    \label{fig:extra_res}
\end{figure*}

\section{Closed-Form Solution Proof}
\label{app:proof}
We aim to minimize the loss function presented in \autoref{eq:loss}, which is
\begin{align*}
    L({\W'}_K, {\W'}_V) =
    \sum_{i=1}^{l}
    \left\lVert {\W'}_K \cc_i - \kk^*_i \right\rVert_2^2
    + \lambda \left\lVert {\W'}_K - {\W}_K \right\rVert_F^2 + \sum_{i=1}^{l}
    \left\lVert {\W'}_V \cc_i - \vv^*_i \right\rVert_2^2
    + \lambda \left\lVert {\W'}_V - {\W}_V \right\rVert_F^2.
\end{align*}
To find the optimal ${\W'}_K$, we differentiate w.r.t.\ it and set to zero:
\begin{align*}
\frac{\partial L({\W'}_K, {\W'}_V)}{\partial {\W'}_K} & = &
\sum_{i=1}^{l}
2 \left( {\W'}_K \cc_i - \kk^*_i \right) \cc_i^\top
+ 2 \lambda \left( {\W'}_K - {\W}_K \right) & = 0 \\
& \Rightarrow &
\sum_{i=1}^{l}
\left( {\W'}_K \cc_i - \kk^*_i \right) \cc_i^\top
+ \lambda \left( {\W'}_K - {\W}_K \right) & = 0 \\
& \Rightarrow &
\sum_{i=1}^{l} {\W'}_K \cc_i \cc_i^\top - \sum_{i=1}^{l} \kk^*_i \cc_i^\top
+ \lambda {\W'}_K - \lambda {\W}_K & = 0 \\
& \Rightarrow &
\sum_{i=1}^{l} {\W'}_K \cc_i \cc_i^\top 
+ \lambda {\W'}_K  & = \sum_{i=1}^{l} \kk^*_i \cc_i^\top + \lambda {\W}_K \\
& \Rightarrow &
\lambda {\W'}_K + \sum_{i=1}^{l} {\W'}_K \cc_i \cc_i^\top  & = \lambda {\W}_K + \sum_{i=1}^{l} \kk^*_i \cc_i^\top \\
& \Rightarrow &
{\W'}_K \left( \lambda \I + \sum_{i=1}^{l} \cc_i \cc_i^\top \right)  & = \lambda {\W}_K + \sum_{i=1}^{l} \kk^*_i \cc_i^\top \\
& \Rightarrow &
{\W'}_K  & = \left( \lambda {\W}_K + \sum_{i=1}^{l} \kk^*_i \cc_i^\top \right) \left( \lambda \I + \sum_{i=1}^{l} \cc_i \cc_i^\top \right)^{-1}.
\end{align*}
The last implication holds because $\cc_i \cc_i^\top$ are symmetric rank-one matrices with a positive eigenvalue and therefore positive semi-definite, 
and $\lambda \I$ is positive definite ($\lambda>0$), which makes their total sum positive definite and therefore invertible.
This makes the obtained solution unique and well-defined.
Similarly, we find the optimal ${\W'}_V$ using the same method and obtain
\begin{align*}
{\W'}_V  & = \left( \lambda {\W}_V + \sum_{i=1}^{l} \vv^*_i \cc_i^\top \right) \left( \lambda \I + \sum_{i=1}^{l} \cc_i \cc_i^\top \right)^{-1},
\end{align*}
thus completing the proof. \textit{Q.E.D.}

\section{Implementation Details}
\label{app:implementation}
We use Stable Diffusion~\cite{rombach2022high} version 1.4 with its default hyperparameters: $50$ diffusion timesteps, a classifier-free guidance~\cite{ho2021classifier} scale of $7.5$, and a maximum number of tokens of $77$. The model generates images of size $512\times512$ pixels.
Unless specified otherwise, we use $\lambda=0.1$ for TIME.
In addition, we apply three simple augmentations to the input source and destination text prompts, $s$ and $d$ respectively.
The augmentations map $s$ and $d$ into:
(i) ``A photo of $[s]$'' and ``A photo of $[d]$''; 
(ii) ``An image of $[s]$'' and ``An image of $[d]$''; and 
(iii) ``A picture of $[s]$'' and ``A picture of $[d]$'', respectively.
The original $s$ and $d$ and their augmentations constitute four lists of corresponding token embeddings $\{\cc_i\}_{i=1}^{l}$, $\{\cc^*_i\}_{i=1}^{l}$ (as denoted in \autoref{sec:method}). We concatenate these lists into a unified corresponding embedding list $\{\cc_i\}_{i=1}^{L}$, $\{\cc^*_i\}_{i=1}^{L}$ and use it for the loss function in \autoref{eq:loss} and its solution in \autoref{eq:closed-form}.

To quantify efficacy, generality, and specificity, we use the CLIP~\cite{clip} ViT-B/32 model as a zero-shot text-based classifier.
When calculating metrics over the MS-COCO~\cite{lin2014microsoft} dataset, we follow standard practice~\cite{rombach2022high, saharia2022photorealistic, ramesh2022hierarchical, balaji2022ediffi}:
We randomly sample $30000$ captions from MS-COCO and generate images based on them.
To ensure a comprehensive evaluation of TIME, we apply each of the $104$ edits in the filtered TIMED independently.
Then, we generate images for $289$ captions with each edited model (except for one with $233$ captions).
Finally, we compute CLIP Score~\cite{hessel2021clipscore} against the $30000$ captions, and FID~\cite{fid} against the entire MS-COCO validation set (center cropped and resized to $512\times512$ pixels).

Our source code and datasets are available in the supplementary material, and we will make them public upon acceptance.

\section{Filtering TIMED for Quantitative Evaluation}
\label{app:dataset_filter}

The goal of this work is to edit implicit assumptions in a text-to-image diffusion model, under the premise that the model has the ability to generate the desired image distribution.
TIME edits the model to promote the generation of the desired image distribution for the requested source prompt.
Note that TIME, whose input does not contain images, is not designed to teach the model new visual concepts, but rather edit the existing implicit assumptions.

Therefore, we check whether the base unedited diffusion model is able to generate the desired image distribution when provided with a prompt that specifies the desired attribute.
In most cases, text-to-image diffusion models are successful in generating images with novel concept compositions.
However, when they fail to do so, model editing techniques based on strictly textual data would naturally fail at their task as well.
This failure is attributed to the model's generative capabilities, and would be different for each pre-trained text-to-image model.

We use the pre-trained unedited Stable Diffusion~\cite{rombach2022high} v1.4 model, and generate $24$ images for each positive destination prompt in TIMED (making this setting an \textit{oracle}).
We then use CLIP~\cite{clip} to classify these images as either the source or destination prompt.
Since the destination prompt was explicitly input into the diffusion model, we expect at least $80\%$ of the images to be classified as the destination prompt.
For testing purposes, we filter out TIMED entries where the oracle model obtained less than $80\%$ accuracy. Out of $147$ entries, we discard of $43$ examples where the oracle model fails.
Note that the generative model mostly succeeds at its task, which is why a majority of entries ($104$ out of $147$) are retained.
We then evaluate our method, the unedited model, and the oracle one on these $104$ entries, and the results are summarized in \autoref{tbl:results}.

We provide the TIMED dataset ($147$ test set and $8$ validation set entries) in the supplementary material.
We also provide the filtered $104$-entry test set to allow future work to easily compare results with TIME on Stable Diffusion v1.4.

\begin{table*}
    \centering
    \begin{tabular}{c c | ccc | ccc}
         \toprule
         & & \multicolumn{3}{c|}{\textbf{Optimizing $\W_V$ only}} & \multicolumn{3}{c}{\textbf{Optimizing $\W_V$ and $\W_K$}} \\
         \vspace{-0.7em}&&&&&&& \\
         \textbf{Augmentations} & $\boldsymbol{\lambda}$
         & \textbf{Generality} ($\uparrow$) & \textbf{Specificity} ($\uparrow$) & \textbf{Mean}  ($\uparrow$) & \textbf{Generality} ($\uparrow$) & \textbf{Specificity} ($\uparrow$) & \textbf{Mean} ($\uparrow$) \\
         \midrule
         \multirow{8}{*}{\textbf{No}}
         & $0.01$ & $55.50\%$ & $73.30\%$ & $\underline{63.17\%}$ & $64.60\%$ & $68.00\%$ & $\underline{66.26\%}$ \\
         & $0.1$ & $51.70\%$ & $71.80\%$ & $60.11\%$ & $60.20\%$ & $67.80\%$ & $63.77\%$ \\
         & $1$ & $51.80\%$ & $69.50\%$ & $59.36\%$ & $61.10\%$ & $68.00\%$ & $64.37\%$ \\
         & $10$ & $51.20\%$ & $68.50\%$ & $58.60\%$ & $61.00\%$ & $67.30\%$ & $64.00\%$ \\
         & $100$ & $48.80\%$ & $69.60\%$ & $57.37\%$ & $57.30\%$ & $68.00\%$ & $62.19\%$ \\
         & $1000$ & $44.30\%$ & $68.00\%$ & $53.65\%$ & $46.60\%$ & $67.50\%$ & $55.14\%$ \\
         & $10000$ & $37.10\%$ & $70.60\%$ & $48.64\%$ & $37.00\%$ & $71.60\%$ & $48.79\%$ \\
         & $100000$ & $21.40\%$ & $80.80\%$ & $33.84\%$ & $21.60\%$ & $81.60\%$ & $34.16\%$ \\
         \midrule
         \multirow{8}{*}{\textbf{Yes}}
         & $0.01$ & $55.50\%$ & $64.90\%$ & $59.83\%$ & $65.10\%$ & $62.30\%$ & $63.67\%$ \\
         & $0.1$ & $59.80\%$ & $69.40\%$ & $\underline{64.24\%}$ & $67.80\%$ & $65.40\%$ & $\mathbf{\underline{66.58\%}}$ \\
         & $1$ & {$57.80\%$} & {$68.90\%$} & {$62.86\%$} & $66.70\%$ & $64.50\%$ & $65.58\%$ \\
         & $10$ & $56.30\%$ & $69.20\%$ & $62.09\%$ & $65.90\%$ & $65.10\%$ & $65.50\%$ \\
         & $100$ & $54.80\%$ & $69.80\%$ & $61.40\%$ & $62.50\%$ & $67.20\%$ & $64.76\%$ \\
         & $1000$ & $51.00\%$ & $67.70\%$ & $58.18\%$ & $57.00\%$ & $68.00\%$ & $62.02\%$ \\
         & $10000$ & $46.50\%$ & $67.90\%$ & $55.20\%$ & $49.30\%$ & $66.50\%$ & $56.62\%$ \\
         & $100000$ & $31.60\%$ & $74.90\%$ & $44.45\%$ & $33.10\%$ & $74.50\%$ & $45.84\%$ \\
         \bottomrule
    \end{tabular}
    \caption{Ablation study results. ``Mean'' is the harmonic mean of generality and specificity. The highest mean in each category is \underline{underlined}, and the highest one overall is also \textbf{\underline{in bold}}.}
    \label{tbl:ablation}
\end{table*}

\section{Ablation Study}
\label{app:ablation_study}
To quantify the effect of each element of our method, we conduct an ablation study using the $8$-entry TIMED validation set.
We measure the effect of optimizing only the value projection matrices $\W_V$ versus optimizing both $\W_V$ and $\W_K$.
We also measure the effect of utilizing the textual augmentations detailed in \autoref{app:implementation}.
Finally, we experiment with different $\lambda$ values to traverse the generality--specificity tradeoff.

We evaluate generality and specificity as described in \autoref{sec:results}, and present the ablation study results in \autoref{tbl:ablation}.
We also calculate the harmonic mean of generality and specificity, and use it to choose the best performing option. Thus, the main TIME algorithm discussed in the paper uses text augmentations, optimizes both $\W_V$ and $\W_K$, and uses $\lambda=0.1$.
Note that while this is the best performing option in terms of harmonic mean, other options may exhibit better specificity or generality. Since there is a natural generality--specificity tradeoff, we use the harmonic mean as a heuristic for choosing an optimal point on the tradeoff.
Different model editing applications may benefit from different hyperparameter tuning strategies.
Our closed-form solution becomes numerically unstable for $\lambda<0.01$.
This can be mitigated by optimizing the loss rather than solving it analytically. 
Because this would entail optimization hyperparameter tuning, we consider it out of scope for this work.

\section{Editing Multiple Assumptions}
\label{app:multi}
In order to edit multiple assumptions in bulk, we can use a natural extension of \autoref{eq:loss} and its corresponding solution in \autoref{eq:closed-form}: sum over all requested edits in both \autoref{eq:loss} and \autoref{eq:closed-form}.
To test this method, we use $82$ assumptions from TIMED (after filtering for the appropriate Stable Diffusion version and removing assumptions with the same source text), and apply the multiple edits version of TIME with $\lambda = 1000$ and $24$ random seeds.
This method proves successful in applying the requested edits, with $89\%$ efficacy and $75\%$ generality.
However, it exhibits low specificity ($47\%$).
We hope and anticipate that future work can mitigate this issue, and provide tools for editing multiple assumptions in bulk without compromising on either generality or specificity.

\section{Comparison to Text Encoder Finetuning}
\label{app:comp-finetune}
As we mention in the main paper, finetuning a neural network has been found to lead to catastrophic forgetting and a drop in performance in general~\cite{mccloskey1989catastrophic, kemker2018measuring}, as well as in the case of model editing~\cite{zhu2020modifying}.
Here, we demonstrate this phenomenon by finetuning the text encoder to map the requested context vectors $\cc_i$ to their target keys $\kk^*_i$ and values $\vv^*_i$. In other words, we optimize the text encoder's weights for the loss function in \autoref{eq:loss} with $\lambda = 0$.
We use the Adam~\cite{kingma2015adam} optimizer for $4000$ iterations with learning rate $0.01$. To achieve a regularization effect over the text encoder parameters, we use weight decay $\eta$.
We run the experiment for different values of $\eta$ and present our results in \autoref{tbl:text_enc_bl}.
In addition to taking significantly more time ($10$ minutes instead of a fraction of a second), finetuning the text encoder fails to achieve a good tradeoff between generality and specificity.
Moreover, when visually examining the generation outputs after finetuning, we often find incoherent images (that do not look realistic) as a result of the catastrophic forgetting property of finetuning.

\begin{table*}
    \centering
    \begin{tabular}{c | ccc}
         \toprule
         $\boldsymbol{\eta}$
         & \textbf{Generality} ($\uparrow$) & \textbf{Specificity} ($\uparrow$) & \textbf{Mean} ($\uparrow$) \\ 
         \midrule
         $1e-6$ & $67.08\%$ & $37.70\%$ & $48.27\%$ \\                  
         $1e-4$ & $54.38\%$ & $41.46\%$ & $47.05\%$ \\                        
         $1e-2$ & $57.92\%$ & $47.40\%$ & $\underline{52.13\%}$ \\  
         $1$    & $71.25\%$ & $35.73\%$ & $47.59\%$ \\                
         $1e2$  & $73.85\%$ & $20.63\%$ & $32.25\%$ \\                   
         $1e4$  & $67.60\%$ & $21.77\%$ & $32.93\%$ \\                    
         $1e6$  & $47.50\%$ & $55.83\%$ & $51.33\%$ \\                
         \midrule
         TIME & $67.80\%$ & $65.40\%$ & $\mathbf{66.58\%}$ \\ 
         \bottomrule
    \end{tabular}
    \caption{Comparison of TIME with finetuning the text encoder for different values of weight decay ($\eta$). ``Mean'' is the harmonic mean of generality and specificity. The highest mean is \textbf{in bold}, and the second-highest is \underline{underlined}.}
    \label{tbl:text_enc_bl}
\end{table*}

\section{Gender Bias Mitigation}
\subsection{Dataset}
\label{app:gender_dataset}
\begin{table}
\begin{center}
\hspace*{-7pt}
\begin{tabular}{c|ll}
\toprule
\textbf{} & \textbf{Source} & \textbf{Destination} \\
\midrule
 \textbf{Editing} & A nurse & A male nurse \\
 \midrule
 \textbf{Validation} & \multicolumn{2}{c}{A photo of a nurse} \\
\midrule
\multirow{5}{*}{
\textbf{Testing}
}
& \multicolumn{2}{c}{A painting of a nurse} \\
& \multicolumn{2}{c}{A nurse working} \\ 
& \multicolumn{2}{c}{A nurse laughing}	\\
& \multicolumn{2}{c}{A nurse in the workplace} \\ 
& \multicolumn{2}{c}{A nurse digital art} \\
\bottomrule
\end{tabular}
\vspace*{-0.15cm}
\end{center}
\caption{A sample of the data used for gender debiasing in professions. The destination prompt is chosen according to the stereotype of the profession (nurse is stereotypically female).}
\label{tbl:gender_dataset}
\vspace*{-0.05cm}
\end{table}

In \autoref{tbl:gender_dataset}, we present a sample of the data used to perform and evaluate TIME for gender debiasing. The professions are taken from the list of stereotypical professions by \cite{zhao-etal-2018-gender}. Some of the stereotypes listed in the original list did not align with the stereotypes observed on the tested text-to-image model (\textit{e.g.}, tailor was listed as stereotypically female, but the model generated a majority of male tailors). Thus we aligned the stereotypes with what is observed in the model.
Moreover, we dropped professions for which the model did not generate pictures of humans (\textit{e.g.}, editor, accountant), and professions for which CLIP was not able to classify the images as male or female (specifically, the profession ``mover''). The dataset is provided in the supplementary material.

\subsection{Implementation Details and Results}
\label{app:gender_implementation}

TIME edits according to the \textit{editing} prompt (from \autoref{tbl:gender_dataset}), without utilizing textual augmentations.
We search for an ideal  $\lambda_p$ per profession, for which $\Delta_p < 0.1$ on the \textit{validation} prompt.

In \autoref{tbl:gender_full_results} we present the full results for every profession we operated on using the \textit{testing} prompts, including the $\lambda_p$ we used to get these results.
Our results are computed across 24 seeds. For computing $\Delta_p$ (as well as $\Delta$ in \autoref{tbl:gender_results}), a distribution of images is required, thus we compute $\Delta_p$ on 8 seeds, and repeat the experiment 3 times to get an average $\Delta_p$.
Note that this is different from computing $\Delta_p$ on each seed and averaging, since the metric $\Delta_p$ is not defined over a single generated image.
To compute the percentage of females in each profession, $F_p$, we use all of the 24 seeds.

\begin{table}
\centering
\begin{tabular}{l|ll|lll}
\toprule
\multirow{2}{*}{\textbf{Profession}} & \multicolumn{2}{c|}{\textbf{Baseline}} & \multicolumn{3}{c}{\textbf{TIME}} \\
\cmidrule{2-6}
 &  $F_p$ &  $\Delta_p$ ($\downarrow$) & $\lambda_p$ &  $F_p$ &  $\Delta_p$ ($\downarrow$) \\
\midrule
                CEO &        04.0\%  &  0.93  \footnotesize{$\pm 0.02$} &  60000 &      35.2\% &  0.30 \footnotesize{$\pm 0.15$ } \\ 
            Analyst &       16.8\%  &  0.67  \footnotesize{$\pm 0.08$} & 160000 &      31.2\% &  0.37 \footnotesize{$\pm 0.12$ } \\
          Assistant &       56.8\%  &  0.20  \footnotesize{$\pm 0.07$} & 250000 &      46.4\% &  0.12 \footnotesize{$\pm 0.13$ } \\
          Attendant &       37.6\%  &  0.30  \footnotesize{$\pm 0.19$} & 120000 &      52.8\% &  0.10 \footnotesize{$\pm 0.08$ } \\
              Baker &       47.2\%  &  0.13  \footnotesize{$\pm 0.05$} & 500000 &      42.4\% &  0.17 \footnotesize{$\pm 0.10$ } \\
          Carpenter &        08.8\%  &  0.82  \footnotesize{$\pm 0.02$} &   8000 &      54.4\% &  0.18 \footnotesize{$\pm 0.06$ } \\
            Cashier &       88.0\%  &  0.75  \footnotesize{$\pm 0.11$} &   1000 &      40.8\% &  0.17 \footnotesize{$\pm 0.10$ } \\
            Cleaner &       70.4\%  &  0.40  \footnotesize{$\pm 0.15$} &  10000 &      43.2\% &  0.12 \footnotesize{$\pm 0.02$ } \\
              Clerk &       43.2\%  &  0.17  \footnotesize{$\pm 0.10$} & 1000000 &     36.8\% &  0.30 \footnotesize{$\pm 0.12$ } \\
Construction worker &       01.6\%  &  0.97  \footnotesize{$\pm 0.02$} &  17000 &      11.2\% &  0.78 \footnotesize{$\pm 0.09$ } \\
               Cook &       42.4\%  &  0.17  \footnotesize{$\pm 0.06$} & 100000 &      66.4\% &  0.32 \footnotesize{$\pm 0.09$ } \\
          Counselor &       55.2\%  &  0.10  \footnotesize{$\pm 0.07$} & 200000 &      34.4\% &  0.32 \footnotesize{$\pm 0.16$ } \\
           Designer &       52.0\%  &  0.12  \footnotesize{$\pm 0.06$} & 150000 &      33.6\% &  0.30 \footnotesize{$\pm 0.11$ } \\
          Developer &       26.4\%  &  0.45  \footnotesize{$\pm 0.11$} &  40000 &      38.4\% &  0.22 \footnotesize{$\pm 0.15$ } \\
             Driver &       16.0\%  &  0.68  \footnotesize{$\pm 0.06$} & 100000 &      31.2\% &  0.42 \footnotesize{$\pm 0.15$ } \\
             Farmer &       02.4\%  &  0.95  \footnotesize{$\pm 0.04$} &  20000 &      49.6\% &  0.12 \footnotesize{$\pm 0.02$ } \\
              Guard &       18.4\%  &  0.62  \footnotesize{$\pm 0.02$} &    100 &      56.8\% &  0.27 \footnotesize{$\pm 0.06$ } \\
        Hairdresser &       72.0\%  &  0.42  \footnotesize{$\pm 0.18$} & 150000 &      53.6\% &  0.10 \footnotesize{$\pm 0.07$ } \\
        Housekeeper &       99.2\%  &  0.98  \footnotesize{$\pm 0.02$} & 0.010 &       56.0\% &  0.13 \footnotesize{$\pm 0.05$ } \\
            Janitor &       41.6\%  &  0.18  \footnotesize{$\pm 0.09$} & 100000 &      56.0\% &  0.15 \footnotesize{$\pm 0.11$ } \\
            Laborer &       01.6\%  &  0.97  \footnotesize{$\pm 0.02$} &   5500 &      42.4\% &  0.15 \footnotesize{$\pm 0.07$ } \\
             Lawyer &       28.8\%  &  0.43  \footnotesize{$\pm 0.16$} & 100000 &      61.6\% &  0.23 \footnotesize{$\pm 0.15$ } \\
          Librarian &       90.4\%  &  0.83  \footnotesize{$\pm 0.06$} &  90000 &      49.6\% &  0.07 \footnotesize{$\pm 0.02$ } \\
            Manager &       22.4\%  &  0.55  \footnotesize{$\pm 0.04$} & 120000 &      35.2\% &  0.32 \footnotesize{$\pm 0.21$ } \\
           Mechanic &       06.4\%  &  0.88  \footnotesize{$\pm 0.08$} &  40000 &      28.8\% &  0.43 \footnotesize{$\pm 0.16$ } \\
              Nurse &      100.0\%  &  1.00  \footnotesize{$\pm 0.00$} &  30000 &      92.0\% &  0.83 \footnotesize{$\pm 0.05$ } \\
          Physician &       12.0\%  &  0.75  \footnotesize{$\pm 0.15$} &  75000 &      40.8\% &  0.23 \footnotesize{$\pm 0.13$ } \\
       Receptionist &       97.6\%  &  0.95  \footnotesize{$\pm 0.04$} &    100 &      58.4\% &  0.20 \footnotesize{$\pm 0.15$ } \\
        Salesperson &       20.0\%  &  0.60  \footnotesize{$\pm 0.23$} & 250000 &      31.2\% &  0.38 \footnotesize{$\pm 0.22$ } \\
          Secretary &       96.8\%  &  0.93  \footnotesize{$\pm 0.05$} &  12500 &      76.7\% &  0.53 \footnotesize{$\pm 0.19$ } \\
            Sheriff &       15.2\%  &  0.73  \footnotesize{$\pm 0.06$} &  43000 &      29.6\% &  0.43 \footnotesize{$\pm 0.05$ } \\
         Supervisor &       47.2\%  &  0.08  \footnotesize{$\pm 0.08$} & 100000 &      62.4\% &  0.25 \footnotesize{$\pm 0.11$ } \\
             Tailor &       25.6\%  &  0.47  \footnotesize{$\pm 0.15$} &  50000 &      70.4\% &  0.42 \footnotesize{$\pm 0.09$ } \\
            Teacher &       84.8\%  &  0.72  \footnotesize{$\pm 0.10$} &  25000 &      35.2\% &  0.32 \footnotesize{$\pm 0.08$ } \\
             Writer &       59.2\%  &  0.22  \footnotesize{$\pm 0.12$} & 125000 &      48.0\% &  0.07 \footnotesize{$\pm 0.02$ } \\
\bottomrule
\end{tabular}
\caption{Full results for gender debiasing of profession, before and after applying TIME.
$F_p$ and $\Delta_p$ are calculated over the testing prompts, which are unseen during editing.
}
\label{tbl:gender_full_results}
\end{table}

\subsection{Variance Across Prompts}
\label{app:gender_different_prompts}

The choice of prompt has a strong effect over the amount of female figures generated from this prompt. For example, the prompt ``A painting of a baker'' produces 36\% females, while the prompt ``a baker in the workplace'' produces 76\%. Moreover, the prompt ``A painting of a designer'' produces 76\% female figures while the prompt ``a designer laughing'' produces 16\%. We observed the phenomenon across different professions. This might hint on the model's training data, that might be biased in different contexts. We leave this to further investigation in future work.

\end{document}